\documentclass[11pt]{article}
\usepackage{acl2013}
\usepackage{times}
\usepackage{url}
\usepackage{latexsym}
\usepackage{graphicx} % more modern
\usepackage{subfigure}

\usepackage{algorithm}
\usepackage{algorithmic}
\usepackage{enumerate}
\usepackage{multirow}
\usepackage{wrapfig}
\usepackage{graphicx}
\usepackage{subfigure}
\usepackage{url}
\usepackage{makecell}
\usepackage{amsmath}
\usepackage{amssymb}

\newcommand{\xv}{\mathbf{x}}

\newcommand{\yv}{\mathbf{y}}

\newcommand{\zv}{\mathbf{z}}
\newcommand{\zvbar}{\bar{\zv}}
\newcommand{\Zv}{\mathbf{Z}}

\newcommand{\wv}{\mathbf{w}}
\newcommand{\Wv}{\mathbf{W}}

\newcommand{\risk}{\mathcal{R}}
\newcommand{\etav}{\boldsymbol \eta}
\newcommand{\muv}{\boldsymbol \mu}

\newcommand{\alphav}{\boldsymbol \alpha}
\newcommand{\betav}{\boldsymbol \beta}
\newcommand{\thetav}{\boldsymbol \theta}
\newcommand{\Thetav}{\boldsymbol \Theta}

\newcommand{\lambdav}{\boldsymbol \lambda }

\newcommand{\Phiv}{\boldsymbol \Phi }

\newcommand{\ep}{\mathbb{E}}

\def\indicator{{\mathbb I}}

\newtheorem{theorem}{\bf{Theorem}}
\newtheorem{definition}[theorem]{\bf{Definition}}

\title{Improved Bayesian Logistic Supervised Topic Models \\ with Data Augmentation}

\author{Jun Zhu,~Xun Zheng,~Bo Zhang \\
  Department of Computer Science and Technology\\
TNLIST Lab and State Key Lab of Intelligent Technology and Systems \\
  Tsinghua University, Beijing, China \\
  {\tt \{dcszj,dcszb\}@tsinghua.edu.cn;~vforveri.zheng@gmail.com} \\
  }

\date{}

\begin{document}
\maketitle
\begin{abstract}
Supervised topic models with a logistic likelihood have two issues that potentially limit their practical use: 1) response variables are usually over-weighted by document word counts; and 2) existing variational inference methods make strict mean-field assumptions. We address these issues by: 1) introducing a regularization constant to better balance the two parts based on an optimization formulation of Bayesian inference; and 2) developing a simple Gibbs sampling algorithm by introducing auxiliary Polya-Gamma variables and collapsing out Dirichlet variables. Our augment-and-collapse sampling algorithm has analytical forms of each conditional distribution without making any restricting assumptions and can be easily parallelized. Empirical results demonstrate significant improvements on prediction performance and time efficiency.
\end{abstract}

\section{Introduction}
As widely adopted in supervised latent Dirichlet allocation (sLDA) models~\cite{Blei:sLDA10,Wang:sLDA09}, one way to improve the predictive power of LDA is to define a likelihood model for the widely available document-level response variables, in addition to the likelihood model for document words. For example, the logistic likelihood model is commonly used
for binary or multinomial responses. By imposing some priors, posterior inference is done with the Bayes' rule. Though powerful, one issue that could limit the use of existing
logistic supervised LDA models is that they treat the document-level response variable as one additional word via a normalized likelihood model.
Although some special treatment is carried out on defining the likelihood of the single response variable, it is normally of a much smaller scale
than the likelihood of the usually tens or hundreds of words in each document. As noted by~\cite{Halpern:2012} and observed in our experiments, this model imbalance
could result in a weak influence of response variables on the topic representations and thus non-satisfactory prediction performance.
Another difficulty arises when dealing with categorical response variables is that the commonly used normal priors are no longer conjugate to the logistic likelihood
and thus lead to hard inference problems. Existing approaches rely on variational approximation techniques which normally make strict mean-field assumptions.

To address the above issues, we present two improvements. First, we present a general framework of Bayesian logistic supervised topic models with a regularization parameter to better balance response variables and words. Technically, instead of doing standard Bayesian inference via Bayes' rule, which requires a normalized likelihood model, we propose to do regularized Bayesian inference~\cite{Zhu:nips11,jun:arXiv13} via solving an optimization problem, where the posterior regularization is defined as an expectation of a logistic loss, a surrogate loss of the expected misclassification error; and a regularization parameter is introduced to balance the surrogate classification loss (i.e., the response log-likelihood) and the word likelihood. The general formulation subsumes standard sLDA as a special case. % when the regularization parameter is set at $1$.

Second, to solve the intractable posterior inference problem of the generalized Bayesian logistic supervised topic models, we present a simple Gibbs sampling algorithm by exploring the ideas of data augmentation~\cite{Tanner:1987,DykMeng2001,Holmes:BA06}. More specifically, we extend Polson's method for Bayesian logistic regression~\cite{Polson:arXiv12} to the generalized logistic supervised topic models, which are much more challenging due to the presence of non-trivial latent variables. Technically, we introduce a set of Polya-Gamma variables, one per document, to reformulate the generalized logistic pseudo-likelihood model (with the regularization parameter) as a scale mixture, where the mixture component is conditionally normal for classifier parameters. Then, we develop a simple and efficient Gibbs sampling algorithms with analytic conditional distributions without Metropolis-Hastings accept/reject steps. For Bayesian LDA models, we can also explore the conjugacy of the Dirichlet-Multinomial prior-likelihood pairs to collapse out the Dirichlet variables (i.e., topics and mixing proportions) to do collapsed Gibbs sampling, which can have better mixing rates~\cite{Griffiths:04}. Finally, our empirical results on real data sets demonstrate significant improvements on time efficiency. The classification performance is also significantly improved by using appropriate regularization parameters. We also provide a parallel implementation with GraphLab~\cite{Gonzalez+al:osdi2012}, which shows great promise in our preliminary studies.

The paper is structured as follows. Sec.~\ref{section:sLdA} introduces logistic supervised topic models as a general optimization problem. Sec.~\ref{section:GibbsSLDA} presents Gibbs sampling algorithms with data augmentation. Sec.~\ref{section:experiments} presents experiments. Sec.~\ref{section:conclusions} concludes.

\section{Logistic Supervised Topic Models}\label{section:sLdA}

We now present the generalized Bayesian logistic supervised topic models.

\subsection{The Generalized Models}
We consider binary classification with a training set $\mathcal{D} = \{ (\wv_d, y_d) \}_{d=1}^{D}$, where the response variable $Y$ takes values from the output space $\mathcal{Y}=\{0, 1\}$. A logistic supervised topic model consists of two parts --- an LDA model~\cite{Blei:03} for describing the words $\Wv = \{\wv_d\}_{d=1}^D$,~where $\wv_d=\{w_{dn}\}_{n=1}^{N_d}$ denote the words within document $d$, and a logistic classifier for considering the supervising signal $\yv =\{ y_d \}_{d=1}^D$. Below, we introduce each of them in turn.

{\bf LDA}: LDA is a hierarchical Bayesian model that posits each document as an admixture of $K$ topics, where each topic $\Phiv_k$ is a multinomial distribution over a $V$-word vocabulary. For document $d$, the generating process is%\\[-.7cm]
\begin{enumerate}
\item draw a topic proportion $\thetav_d \sim \mathrm{Dir}(\alphav)$\\[-.65cm]
\item for each word $n = 1,2, \dots, N_d$:\\[-.65cm]
\begin{enumerate}
\item draw a topic\footnote{A $K$-binary vector with only one entry equaling to 1.} $z_{dn} \sim \mathrm{Mult}(\thetav_d)$\\[-.5cm]
\item draw the word $w_{dn} \sim \mathrm{Mult}(\Phiv_{z_{dn}})$\\[-.7cm]
\end{enumerate}
\end{enumerate}
%its topic proportion $\thetav_d$ is a multinomial distribution drawn from a Dirichlet prior $\mathrm{Dir}(\alphav)$. Let $\wv_d=\{w_{dn}\}_{n=1}^N$ denote the words appearing in document $d$. For the $n$-th word $w_{dn}$, a topic assignment vector\footnote{A $K$-dimension binary vector with only one entry equaling to 1.} $z_{dn}$ is drawn from $\thetav_d$ and $w_{dn}$ is drawn from $\Phiv_{z_{dn}}$. In short, the generative process of $d$ is
%\begin{equation}
%\thetav_d \sim \mathrm{Dir}(\alphav),~ z_{dn} \sim \mathrm{Mult}(\thetav_d),~w_{dn} \sim\mathrm{Mult}(\Phiv_{z_{dn}}),
%\end{equation}
where $\mathrm{Dir}(\cdot)$ is a Dirichlet distribution; $\mathrm{Mult}(\cdot)$ is a multinomial distribution; and $\Phiv_{z_{dn}}$ denotes the topic selected by the non-zero entry of $z_{dn}$. For fully-Bayesian LDA, the topics are random samples from a Dirichlet prior, $\Phiv_k \sim \textrm{Dir}(\betav)$.

Let $\zv_d = \{z_{dn}\}_{n=1}^{N_d}$ denote the set of topic assignments for document $d$. Let $\Zv = \{\zv_d\}_{d=1}^D$ and $\Thetav = \{\thetav_d\}_{d=1}^D$ denote all the topic assignments and mixing proportions for the entire corpus. LDA infers the posterior distribution $ p(\Thetav, \Zv, \Phiv | \Wv) \propto p_0(\Thetav, \Zv, \Phiv) p(\Wv| \Zv, \Phiv)$, where $p_0(\Thetav, \Zv, \Phiv) = \big( \prod_d p(\thetav_d | \alphav) \prod_n p(z_{dn}|\theta_d) \big) \prod_k p(\Phi_k|\betav)$ is the joint distribution defined by the model. As noticed in~\cite{Zhu:nips12}, the posterior distribution by Bayes' rule is equivalent to the solution of an information theoretical optimization problem\\[-1cm]

{\small \setlength\arraycolsep{-1pt} \begin{eqnarray}\label{problem:RegBayesLDA}
&&\min_{q(\Theta, \Zv, \Phiv) } \! \mathrm{KL}( q(\Theta, \Zv, \Phiv) \Vert p_0(\Theta, \Zv, \Phiv) ) \!-\! \ep_q[\log p(\Wv |  \Zv, \Phiv)] \nonumber \\
&&~~~~~\mathrm{s.t.}: q(\Theta, \Zv, \Phiv) \in \mathcal{P},
\end{eqnarray}}\\[-1cm]

\noindent where $\mathrm{KL}(q||p)$ is the Kullback-Leibler divergence from $q$ to $p$ and $\mathcal{P}$ is the space of probability distributions.

%In fact, if we add the constant $\log p(\Wv)$ to the objective, the problem becomes the minimization of $\mathrm{KL}(q(\Theta,\Zv, \Phiv) || p(\Theta,\Zv, \Phiv|\Wv))$. Thus, the optimum solution is $q(\Theta,\Zv, \Phiv) = p(\Theta,\Zv, \Phiv | \Wv)$.

{\bf Logistic classifier}: To consider binary supervising information, a logistic supervised topic model (e.g., sLDA) builds a logistic classifier using the topic representations as input features\\[-1cm]

{\small \begin{eqnarray}
p(y=1|\etav, \zv) = \frac{\exp(\etav^\top \zvbar)}{1 + \exp(\etav^\top \zvbar)},
\end{eqnarray}}\\[-1cm]

\noindent where $\bar{\zv}$ is a $K$-vector with $\bar{z}_k = \frac{1}{N}\sum_{n=1}^{N}\mathbb{I}(z_{n}^k=1)$, and $\mathbb{I}(\cdot)$ is an indicator function that equals to 1 if predicate holds otherwise 0. If the classifier weights $\etav$ and topic assignments $\zv$ are given, the prediction rule is \\[-1cm]

{\small \begin{eqnarray}
\hat{y}|_{\etav,\zv} = \indicator( p(y=1|\etav, \zv) > 0.5 ) = \indicator( \etav^\top \zvbar > 0).
\end{eqnarray}}\\[-1cm]

\noindent Since both $\etav$ and $\Zv$ are hidden variables, we propose to infer a posterior distribution $q(\etav, \Zv)$ that has the minimal expected log-logistic loss\\[-1cm]

{\small \begin{eqnarray}
\risk(q(\etav, \Zv)) = - \sum_d \ep_q[\log p(y_d | \etav, \zv_d)],
\end{eqnarray}} \\[-1cm]

\noindent which is a good surrogate loss for the expected misclassification loss, $\sum_d \ep_q[ \indicator(\hat{y}|_{\etav,\zv_d} \neq y_d) ]$, of a Gibbs classifier that randomly draws a model $\etav$ from the posterior distribution and makes predictions~\cite{McAllester2003,Germain:icml09}. In fact, this choice is motivated from the observation that logistic loss has been widely used as a convex surrogate loss for the misclassification loss~\cite{Rosasco:04} in the task of fully observed binary classification. Also, note that the logistic classifier and the LDA likelihood are coupled by sharing the latent topic assignments $\zv$. The strong coupling makes it possible to learn a posterior distribution that can describe the observed words well and make accurate predictions.

{\bf Regularized Bayesian Inference}: To integrate the above two components for hybrid learning, a logistic supervised topic model solves the joint Bayesian inference problem\\[-1cm]

\setlength\arraycolsep{1pt}  {\small \begin{eqnarray}\label{problem:sLDA}
\min_{q(\etav, \Thetav, \Zv, \Phiv)} && \mathcal{L}(q(\etav, \Thetav, \Zv, \Phiv)) + c \risk(q(\etav, \Zv)) \\ %\ep_p[\log p(\yv | \etav, \Zv)] \\
\textrm{s.t.:} && ~ q(\etav, \Thetav, \Zv, \Phiv) \in \mathcal{P}, \nonumber
\end{eqnarray}}\\[-1cm]

\noindent where $\mathcal{L}(q) \!=\! \mathrm{KL}(q||p_0(\etav, \Thetav, \Zv, \Phiv)) \!-\! \ep_q[\log p(\Wv | \Zv, \Phiv)]$ is the objective for doing standard Bayesian inference with the classifier weights $\etav$; $p_0(\etav, \Thetav, \Zv, \Phiv) = p_0(\etav) p_0(\Thetav, \Zv, \Phiv)$; and $c$ is a regularization parameter balancing the influence from response variables and words.

In general, we define the pseudo-likelihood for the supervision information\\[-1cm]

{\small \begin{eqnarray}\label{eq:pseudo-likelihood}
\psi(y_d|\zv_d, \etav) =  p^c(y_d|\etav, \zv_d) = \frac{ \{\exp(\etav^\top \zvbar_d) \}^{c y_d} }{ (1 + \exp(\etav^\top \zvbar_d))^c},
\end{eqnarray}}\\[-1cm]

\noindent which is un-normalized if $c \neq 1$. But, as we shall see this un-normalization does not affect our subsequent inference. Then, the generalized inference problem~(\ref{problem:sLDA}) of logistic supervised topic models can be written in the ``standard" Bayesian inference form~(\ref{problem:RegBayesLDA})\\[-1cm]

{\small \begin{eqnarray}\label{problem:sLDA2}
\min_{q(\etav, \Thetav, \Zv, \Phiv)} && \mathcal{L}(q(\etav, \Thetav, \Zv, \Phiv)) - \ep_q[ \log \psi(\yv | \Zv, \etav) ] \\ %\ep_p[\log p(\yv | \etav, \Zv)] \\
\textrm{s.t.:} && ~ q(\etav, \Thetav, \Zv, \Phiv) \in \mathcal{P}, \nonumber
\end{eqnarray}}\\[-1cm]

\noindent where $\psi(\yv|\Zv, \etav) = \prod_d  \psi(y_d|\zv_d, \etav)$. It is easy to show that the optimum solution of problem (\ref{problem:sLDA}) or the equivalent problem (\ref{problem:sLDA2}) is the posterior distribution with supervising information, i.e.,\\[-1cm]

{\small \begin{eqnarray}
q(\etav, \Thetav, \Zv, \Phiv) = \frac{p_0(\etav, \Thetav, \Zv, \Phiv) p(\Wv|\Zv,\Phiv) \psi(\yv|\etav, \Zv)}{\phi(\yv, \Wv)}. \nonumber
\end{eqnarray}}\\[-1cm]

\noindent where $\phi(\yv, \Wv)$ is the normalization constant to make $q$ a distribution. We can see that when $c=1$, the model reduces to the standard sLDA, which in practice has the imbalance issue that the response variable (can be viewed as one additional word) is usually dominated by the words. This imbalance was noticed in~\cite{Halpern:2012}. We will see that $c$ can make a big difference later.

{\bf Comparison with MedLDA}: The above formulation of logistic supervised topic models as an instance of regularized Bayesian inference provides a direct comparison with the max-margin supervised topic model (MedLDA)~\cite{Zhu:nips12}, which has the same form of the optimization problems. The difference lies in the posterior regularization, for which MedLDA uses a hinge loss of an expected classifier while the logistic supervised topic model uses an expected log-logistic loss. Gibbs MedLDA~\cite{Zhu:icml13} is another max-margin model that adopts
the expected hinge loss as posterior regularization. As we shall see in the experiments, by using appropriate regularization constants, logistic supervised topic models achieve comparable performance as max-margin methods. We note that the relationship between a logistic loss and a hinge loss has been discussed extensively in various settings~\cite{Rosasco:04,Globerson:EG07}. But the presence of latent variables poses additional challenges in carrying out a formal theoretical analysis of these surrogate losses~\cite{Lin:TR01} in the topic model setting.

%, where the posterior regularization on $p(\etav,\Thetav, \Zv,\Phiv)$ is introduced by the expected log-logistic loss. This interpretation leads to

\vspace{-.1cm}
\subsection{ Variational Approximation Algorithms}
\vspace{-.1cm}

The commonly used normal prior for $\etav$ is non-conjugate to the logistic likelihood, which makes the posterior inference hard. Moreover, the latent variables $\Zv$ make the inference problem harder than that of Bayesian logistic regression models~\cite{Chen:JASA99,Meyer:JASA02,Polson:arXiv12}. Previous algorithms to solve problem~(\ref{problem:sLDA}) rely on variational approximation techniques. It is easy to show that the variational method~\cite{Wang:sLDA09} is a coordinate descent algorithm to solve problem~(\ref{problem:sLDA}) with the additional fully-factorized constraint $q(\etav, \Theta, \Zv, \Phiv) = q(\etav) (\prod_d q(\thetav_d) \prod_n q(z_{dn})) \prod_k q(\Phiv_k)$ and a variational approximation to the expectation of the log-logistic likelihood, which is intractable to compute directly. Note that the non-Bayesian treatment of $\etav$ as unknown parameters in~\cite{Wang:sLDA09} results in an EM algorithm, which still needs to make strict mean-field assumptions together with a variational bound of the expectation of the log-logistic likelihood. In this paper, we consider the full Bayesian treatment, which can principally consider prior distributions and infer the posterior covariance.

%The variational approximation is needed as we do not have an analytical form.

\section{A Gibbs Sampling Algorithm}\label{section:GibbsSLDA}

Now, we present a simple and efficient Gibbs sampling algorithm for the generalized Bayesian logistic supervised topic models. % by exploring the ideas of data augmentation.

\subsection{Formulation with Data Augmentation}

Since the logistic pseudo-likelihood $\psi(\yv|\Zv,\etav)$ is not conjugate with normal priors, it is not easy to derive the sampling algorithms directly. Instead, we develop our algorithms by introducing auxiliary variables, which lead to a scale mixture of Gaussian components and analytic conditional distributions for automatical Bayesian inference without an accept/reject ratio. Our algorithm represents a first attempt to extend Polson's approach~\cite{Polson:arXiv12} to deal with highly non-trivial Bayesian latent variable models. Let us first introduce the Polya-Gamma variables.
\begin{definition} \cite{Polson:arXiv12}~A random variable $X$ has a Polya-Gamma distribution, denoted by $X \!\sim\! \mathcal{PG}(a, b)$, if\\[-1cm]

{\small \begin{eqnarray}
X = \frac{1}{2 \pi^2}\sum_{i=1}^\infty \frac{g_k}{(i-1)^2/2 + b^2/(4\pi^2)}, \nonumber
\end{eqnarray}}\\[-1cm]

\noindent where $a,b>0$ and each $g_i \sim \mathcal{G}(a, 1)$ is an independent Gamma random variable.
\end{definition}
Let $\omega_d = \etav^\top \zvbar_d$. Then, using the ideas of data augmentation~\cite{Tanner:1987,Polson:arXiv12}, we can show that the generalized pseudo-likelihood can be expressed as\\[-1cm]

{\small \setlength\arraycolsep{1pt} \begin{eqnarray}
&& \psi(y_d|\zv_d, \etav) = \frac{1}{2^c} e^{ \kappa_d \omega_d } \! \int_0^\infty \!\! \exp\Big( \! - \frac{\lambda_d \omega_d^2}{2} \Big) p(\lambda_d | c, 0) d \lambda_d, \nonumber
\end{eqnarray}}\\[-1cm]

\noindent where $\kappa_d = c(y_d - 1/2)$ and $\lambda_d$ is a Polya-Gamma variable with parameters $a=c$ and $b=0$.
\iffalse
\begin{proof}
Using the definition in~(\ref{eq:pseudo-likelihood}) and letting $\omega = \etav^\top \zvbar_d$, we can apply Theorem 1 in~\cite{Polson:arXiv12} to finish the proof.
\end{proof}
\fi
This result indicates that the posterior distribution of the generalized Bayesian logistic supervised topic models, i.e., $q(\etav, \Thetav, \Zv, \Phiv)$, can be expressed as the marginal of a higher dimensional distribution that includes the augmented variables $\lambdav$. The complete posterior distribution is\\[-1cm]

{\small \setlength\arraycolsep{1pt}\begin{eqnarray}
q(\etav, \lambdav, \Thetav, \Zv, \Phiv) = \frac{ p_0(\etav, \Thetav, \Zv, \Phiv) p(\Wv | \Zv, \Phiv) \phi(\yv, \lambdav | \Zv, \etav) }{\psi(\yv, \Wv)}, \nonumber
\end{eqnarray}}\\[-1cm]

\noindent where the pseudo-joint distribution of $\yv$ and $\lambdav$ is\\[-1cm]

{\small \begin{eqnarray}
\phi(\yv, \lambdav | \Zv, \etav) = \prod_d \exp\Big( \kappa_d \omega_d - \frac{\lambda_d \omega_d^2}{2} \Big) p(\lambda_d | c, 0). \nonumber
\end{eqnarray}}\\[-1cm]

\iffalse
In fact, we can show that the complete posterior distribution is the solution of augmented problem\\[-1cm]

{\small \setlength\arraycolsep{0pt}\begin{eqnarray}\label{eq:JointInf}
&&\min_{q(\etav, \lambdav, \Thetav, \Zv, \Phiv) } \mathcal{L}(q(\etav, \lambdav, \Thetav, \Zv, \Phiv) ) - \ep_q [ \log \phi(\yv, \lambdav | \Zv, \etav) ], \nonumber
\end{eqnarray}}\\[-1cm]

\noindent which is again subject to the normalization constraint that $q(\etav, \lambdav, \Thetav, \Zv, \Phiv) \in \mathcal{P}$.
\fi

\subsection{Inference with Collapsed Gibbs Sampling}

%With the above data augmentation formulation, we can develop efficient Monte Carlo methods to infer the posterior distribution.

Although we can do Gibbs sampling to infer the complete posterior distribution $q(\etav, \lambdav,\Thetav, \Zv, \Phiv)$ and thus $q(\etav, \Thetav, \Zv, \Phiv)$ by ignoring $\lambdav$, the mixing rate would be slow due to the large sample space. One way to effectively improve mixing rates is to integrate out the intermediate variables $(\Thetav, \Phiv)$ and build a Markov chain whose equilibrium distribution is the marginal distribution $q(\etav, \lambdav, \Zv)$. We propose to use collapsed Gibbs sampling, which has been successfully used in LDA \cite{Griffiths:04}. For our model, the collapsed posterior distribution is\\[-1cm]

{\small \setlength\arraycolsep{1pt} \begin{eqnarray}
 q(\etav, \lambdav, \Zv)  && \propto  p_0(\etav) p(\Wv, \Zv|\alphav, \betav) \phi(\yv, \lambdav|\Zv, \etav)  \nonumber \\
             && = p_0(\etav) \prod_{k=1}^{K}\frac{\delta(\mathbf{C}_k + \betav)}{\delta(\betav)}  \prod_{d=1}^{D}  \Big[ \frac{\delta(\mathbf{C}_d + \alphav)}{\delta(\alphav)}   \nonumber \\
             && ~~~ \times \exp\Big( \kappa_d \omega_d - \frac{\lambda_d \omega_d^2}{2} \Big) p(\lambda_d | c, 0) \Big], \nonumber
\end{eqnarray}}\\[-.9cm]

\noindent where $\delta(\xv)=\frac{\prod_{i=1}^{\mathrm{dim}(\xv)}\Gamma(x_i)}{\Gamma(\sum_{i=1}^{\mathrm{dim}(\xv)} x_i)}$, $C_k^t$ is the number of times the term $t$ being assigned to topic $k$ over the whole corpus and $\mathbf{C}_k=\{C_k^t\}_{t=1}^{V}$; $C_d^k$ is the number of times that terms being associated with topic $k$ within the $d$-th document and $\mathbf{C}_d=\{C_d^k\}_{k=1}^{K}$. Then, the conditional distributions used in collapsed Gibbs sampling are as follows.

{\bf For $\etav$}: for the commonly used isotropic Gaussian prior $p_0(\etav) = \prod_k \mathcal{N}(\eta_k; 0, \nu^2)$, we have\\[-1cm]

{\small \begin{eqnarray}\label{eq:GibbsEta}
q(\etav | \Zv, \lambdav ) && \propto p_0(\etav) \prod_d \exp\Big( \kappa_d \omega_d - \frac{\lambda_d \omega_d^2}{2} \Big) \nonumber \\
%                         &&\propto \exp\big( - \frac{ \Vert \etav \Vert_2^2}{2 \nu^2} - \frac{ \sum_d \lambda_d(\etav^\top \zvbar_d)^2 - 2\kappa_d \etav^\top \zvbar_d }{2}  \big) \nonumber \\
%                         &&\propto \exp\Big( -\sum_k \frac{\eta_k^2}{2 \nu^2} - \sum_d \frac{ c^2 \etav^\top \zvbar_d \zvbar_d^\top \etav - 2 c y_d(\lambda_d + c\ell) \zvbar_d^\top \etav}{2 \lambda_d}  \Big) \nonumber \\
%                         &&= \exp\Big( -\frac{1}{2}\etav^\top ( \frac{1}{\nu^2} I + c^2 \sum_d \frac{\zvbar_d \zvbar_d^\top}{\lambda_d} )\etav + (c\sum_d y_d \frac{\lambda_d + c\ell }{\lambda_d}\zvbar_d )^\top \etav \Big) \nonumber \\
                         &&= \mathcal{N}(\etav; \muv, \Sigma),
\end{eqnarray}}\\[-1.1cm]

\noindent where the posterior mean is $\muv = \Sigma ( \sum_d \kappa_d \zvbar_d )$ and the covariance is $\Sigma = (\frac{1}{\nu^2}I + \sum_d \lambda_d \zvbar_d \zvbar_d^\top)^{-1}$. We can easily draw a sample from a $K$-dimensional multivariate Gaussian distribution. The inverse can be robustly done using Cholesky decomposition, an $O(K^3)$ procedure. Since $K$ is normally not large, the inversion can be done efficiently. %Note that more efficient implementation exists, e.g., by solving a linear equation to replace the inversion.

{\bf For $\Zv$}: The conditional distribution of $\Zv$ is\\[-1cm]

{\small \begin{eqnarray}
q(\Zv | \etav, \lambdav ) && \propto \prod_{k=1}^{K}\frac{\delta(\mathbf{C}_k + \betav)}{\delta(\betav)} \prod_{d=1}^{D} \Big[ \frac{\delta(\mathbf{C}_d + \alphav)}{\delta(\alphav)} \nonumber \\
&& ~~ \times \exp\Big( \kappa_d  \omega_d - \frac{\lambda_d \omega_d^2}{2} \Big) \Big]. \nonumber
\end{eqnarray}}\\[-1cm]

\noindent By canceling common factors, we can derive the local conditional of one variable $z_{dn}$ as:\\[-1cm]

{\small \setlength\arraycolsep{1pt} \begin{eqnarray}\label{eqn:transitionProb}
q(z_{dn}^k = 1 &|& \Zv_{\neg}, \etav, \lambdav, w_{dn}=t ) \nonumber \\
&& \propto \frac{ (C_{k,\neg n}^{t}+\beta_t) (C_{d,\neg n}^{k}+\alpha_k) }{\sum_t C_{k,\neg n}^t + \sum_{t=1}^V \beta_t} \exp\Big( \gamma \kappa_d \eta_k  \nonumber \\
&& ~~  - \lambda_d \frac{\gamma^2 \eta_k^2 + 2 \gamma(1-\gamma)\eta_k \Lambda_{dn}^k }{2} \Big),
\end{eqnarray}}\\[-1cm]

\noindent where $C_{\cdot,\neg n}^{\cdot}$ indicates that term $n$ is excluded from the corresponding document or topic; $\gamma = \frac{1}{N_d}$; and $\Lambda_{dn}^k = \frac{1}{N_d-1} \sum_{k^\prime} \eta_{k^\prime} C_{d, \neg n}^{k^\prime}$ is the discriminant function value without word $n$. We can see that the first term is from the LDA model for observed word counts and the second term is from the supervising signal $\yv$.

\begin{algorithm}[t]
\caption{ for collapsed Gibbs sampling}\label{alg:GibbsAlg}
\begin{algorithmic}[1]
%   \STATE {\bfseries Input:} corpus $\data$, regularization constants $(c, \nu^2, \ell)$ and topic number $K$.
   \STATE {\bfseries Initialization:} set $\lambdav = 1$ and randomly draw $z_{dn}$ from a uniform distribution.
   \FOR{$m=1$ {\bfseries to} $M$}
   \STATE draw a classifier from the distribution~(\ref{eq:GibbsEta})
    \FOR{$d=1$ {\bfseries to} $D$}
        \FOR{each word $n$ in document $d$}
             \STATE draw the topic using distribution~(\ref{eqn:transitionProb})
        \ENDFOR
       \STATE draw $\lambda_d$ from distribution~(\ref{eq:GibbsLambda}).
   \ENDFOR
   \ENDFOR
\end{algorithmic}
\end{algorithm}

{\bf For $\lambdav$}: Finally, the conditional distribution of the augmented variables $\lambdav$ is
{\small \begin{eqnarray}\label{eq:GibbsLambda}
 q(\lambda_d | \Zv, \etav) &&\propto \exp\Big( - \frac{\lambda_d \omega_d^2}{2} \Big) p(\lambda_d | c, 0) \nonumber \\
%                         &&\propto \frac{1}{\sqrt{ 2 \pi \lambda_d }} \exp\big( -\frac{c^2 \zeta_d^2}{2 \lambda_d} - \frac{\lambda_d}{2} \big) \nonumber \\
                         && =  \mathcal{PG}\big(\lambda_d; c, \omega_d  \big),
\end{eqnarray}}
\noindent which is a Polya-Gamma distribution. The equality has been achieved by using the construction definition of the general $\mathcal{PG}(a, b)$ class through an exponential tilting of the $\mathcal{PG}(a,0)$ density~\cite{Polson:arXiv12}. To draw samples from the Polya-Gamma distribution, %a naive implementation of the sampling using the infinite sum-of-Gamma representation is not efficient and it also involves a potentially inaccurate step of truncating the infinite sum. Here
we adopt the efficient method\footnote{The basic sampler was implemented in the R package BayesLogit. We implemented the sampling algorithm in C++ together with our topic model sampler.} proposed in~\cite{Polson:arXiv12}, which draws the samples through drawing samples from the closely related exponentially tilted Jacobi distribution.

With the above conditional distributions, we can construct a Markov chain which iteratively draws samples of $\etav$ using Eq. (\ref{eq:GibbsEta}), $\Zv$ using Eq. (\ref{eqn:transitionProb}) and $\lambdav$ using Eq. (\ref{eq:GibbsLambda}), with an initial condition. In our experiments, we initially set $\lambdav=1$ and randomly draw $\Zv$ from a uniform distribution. In training, we run the Markov chain for $M$ iterations (i.e., the burn-in stage), as outlined in Algorithm~\ref{alg:GibbsAlg}. Then, we draw a sample $\hat{\etav}$ as the final classifier to make predictions on testing data. As we shall see, the Markov chain converges to stable prediction performance with a few burn-in iterations.

%and estimate the classifier $\hat{\etav} = \frac{1}{J} \sum_{j=1}^{J} \etav^{(j)}$. In practice, a single sample (i.e., $J=1$) usually suffices.

%%--------------------------------------------------------
\subsection{Prediction}

To apply the classifier $\hat{\etav}$ on testing data, we need to infer their topic assignments. We take the approach in~\cite{Zhu:jmlr12,Zhu:nips12}, which uses a point estimate of topics $\Phiv$ from training data and makes prediction based on them. Specifically, we use the MAP estimate $\hat{\Phiv}$ to replace the probability distribution $p(\Phiv)$. For the Gibbs sampler, an estimate of $\hat{\Phiv}$ using the samples is $\hat{\phi}_{kt} \propto {C_k^t} + \beta_t$. Then, given a testing document $\wv$, we infer its latent components $\zv$ using $\hat{\Phiv}$ as $p(z_n=k|\mathbf{z}_{\neg n}) \propto \hat{\phi}_{kw_{n}}(C_{\neg n}^{k} + \alpha_k)$, where $C_{\neg n}^{k}$ is the times that the terms in this document $\wv$ assigned to topic $k$ with the $n$-th term excluded.

%We draw $L$ samples $\{\mathbf{Z}^{(i)}\}$ and  estimate estimate the expectation statistics \\[-1.1cm]

%\begin{equation}\label{eqn:E-z-bar-gibbs}
%\ep[ \bar{z}_{dk} ] = \frac{1}{N_d} \sum_{n=1}^{N_d} \ep[ z_{dn} ], ~\text{and}~\ep[ z_{dn} ] = \frac{1}{L}\sum_{i=1}^{L} z_{dn}^{(i)}.
%\end{equation}\\[-1cm]

\section{Experiments} \label{section:experiments}

We present empirical results and sensitivity analysis to demonstrate the efficiency and prediction performance\footnote{Due to space limit, the topic visualization (similar to that of MedLDA) is deferred to a longer version.} of the generalized logistic supervised topic models on the 20Newsgroups (20NG) data set, which contains about 20,000 postings within 20 news groups. We follow the same setting as in~\cite{Zhu:jmlr12} and remove a standard list of stop words for both binary and multi-class classification. For all the experiments, we use the standard normal prior $p_0(\etav)$ (i.e., $\nu^2 = 1$) and the symmetric Dirichlet priors $\alphav  = \frac{\alpha}{K} {\boldsymbol 1},~\betav = 0.01 \times {\boldsymbol 1}$, where ${\boldsymbol 1}$ is a vector with all entries being $1$. For each setting, we report the average performance and the standard deviation with five randomly initialized runs.

%All the experiments are done on a standard desktop computer.

\subsection{Binary classification}
\begin{figure*}\vspace{-.2cm}%[h]
\centering
\subfigure[accuracy]{\includegraphics[height=1.4in]{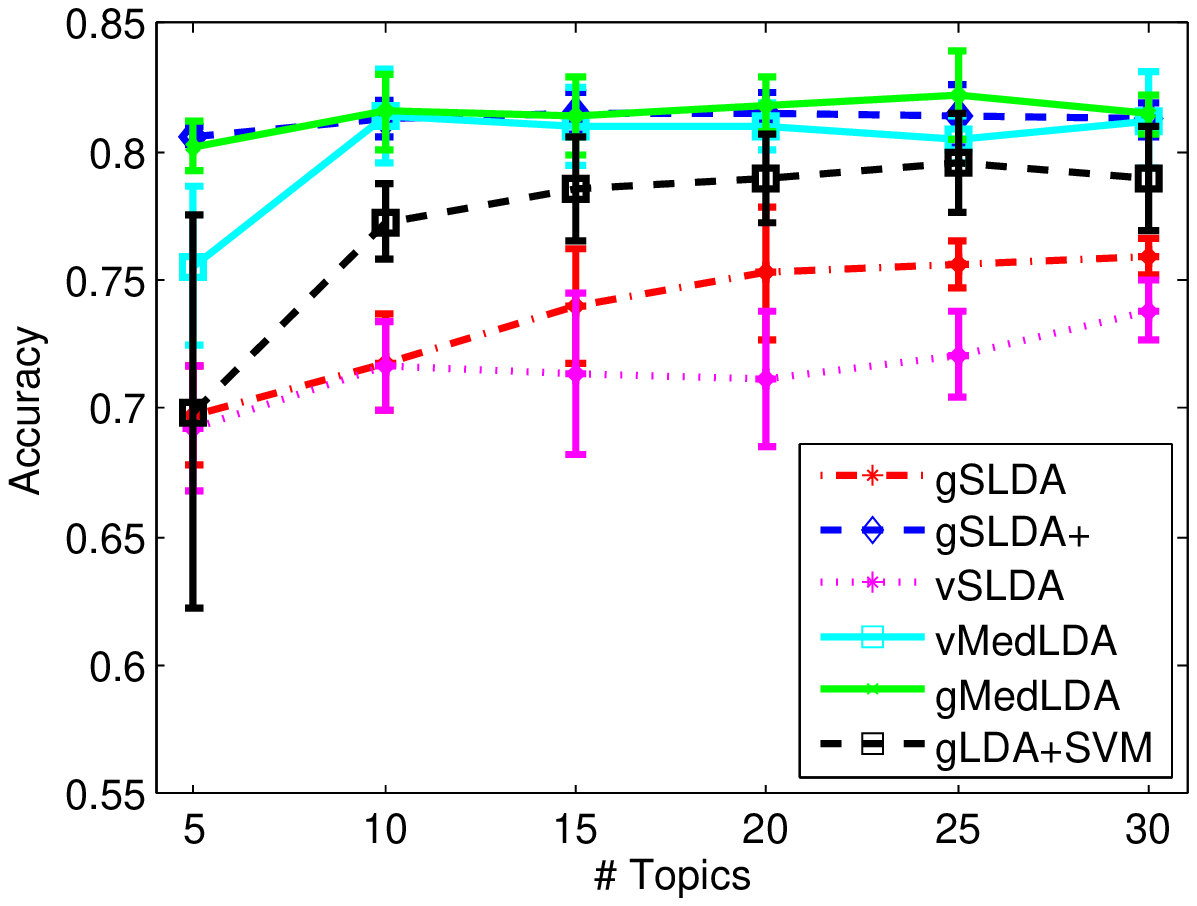}}\hspace{0.4cm}
\subfigure[training time]{\includegraphics[height=1.4in]{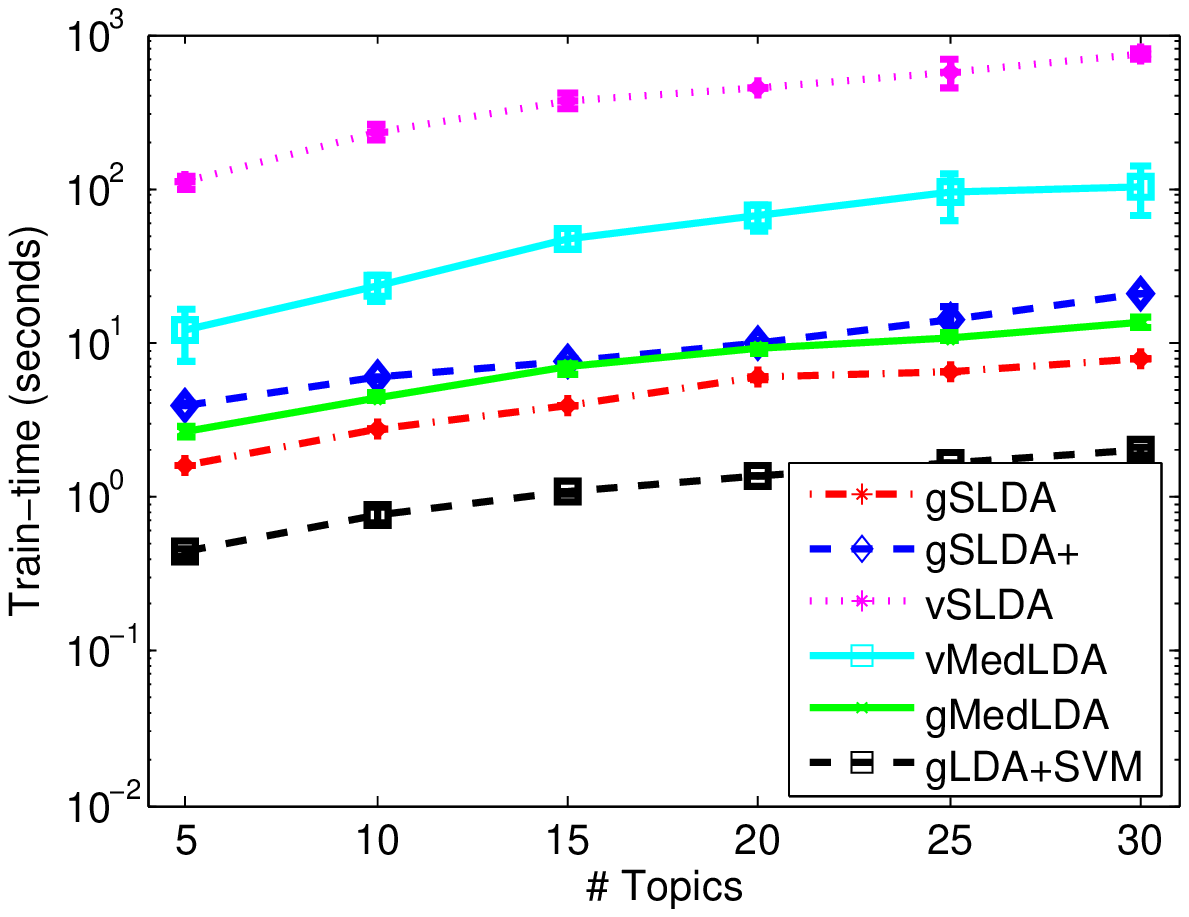}}\hspace{0.4cm}
\subfigure[testing time]{\includegraphics[height=1.4in]{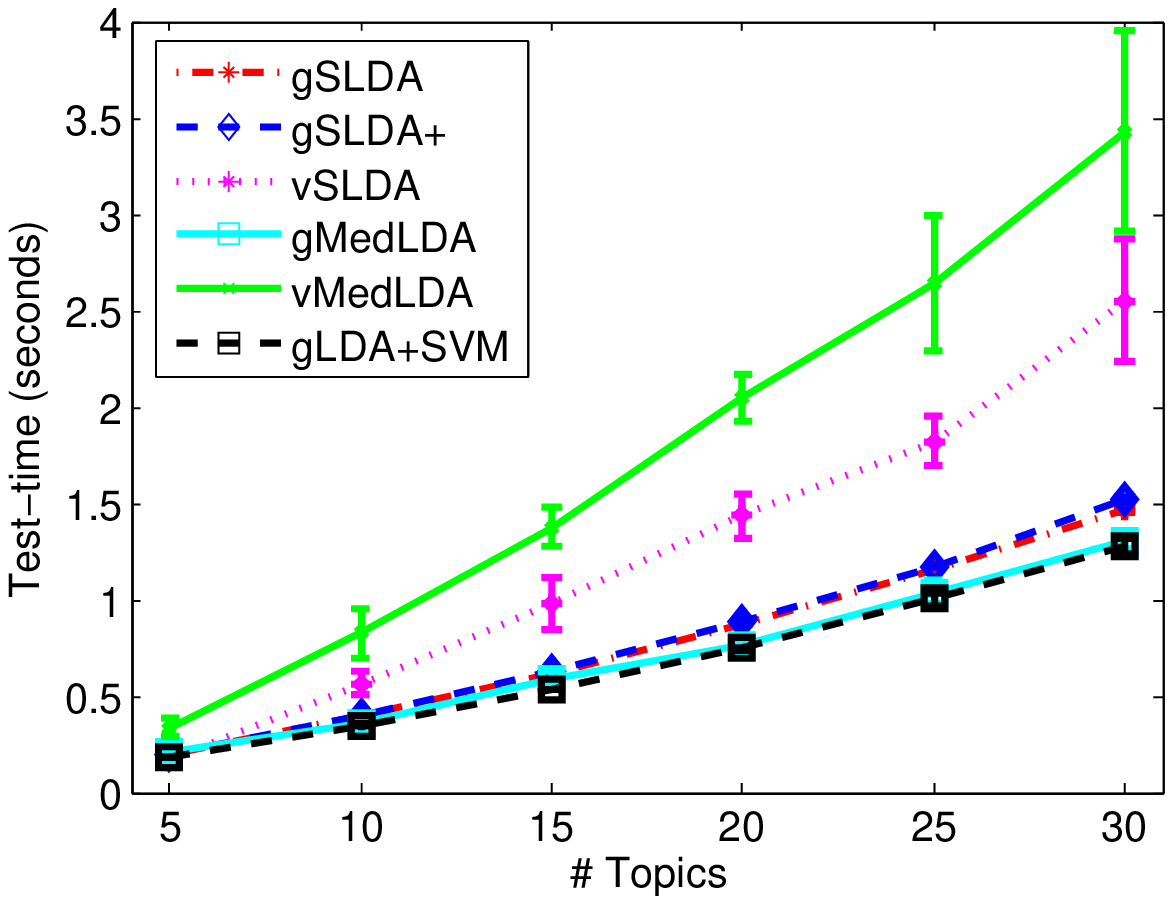}}\vspace{-.3cm}
\caption{Accuracy, training time (in log-scale) and testing time on the 20NG binary data set.}\label{fig:20ngBinary}\vspace{-.3cm}
\end{figure*}

Following the same setting in~\cite{Simon:nips09,Zhu:jmlr12}, the task is to distinguish postings of the newsgroup \emph{alt.atheism} and those of the group \emph{talk.religion.misc}. The training set contains 856 documents and the test set contains 569 documents. We compare the generalized logistic supervised LDA using Gibbs sampling (denoted by gSLDA) with various competitors, including the standard sLDA using variational mean-field methods (denoted by vSLDA)~\cite{Wang:sLDA09}, the MedLDA model using variational mean-field methods (denoted by vMedLDA)~\cite{Zhu:jmlr12}, and the MedLDA model using collapsed Gibbs sampling algorithms (denoted by gMedLDA)~\cite{Zhu:nips12}. We also include the unsupervised LDA using collapsed Gibbs sampling as a baseline, denoted by gLDA. For gLDA, we learn a binary linear SVM on its topic representations using SVMLight~\cite{joachims1999making}. The results of DiscLDA~\cite{Simon:nips09} and linear SVM on raw bag-of-words features were reported in~\cite{Zhu:jmlr12}. For gSLDA, we compare two versions -- the standard sLDA with $c=1$ and the sLDA with a well-tuned $c$ value. To distinguish, we denote the latter by gSLDA+. We set $c=25$ for gSLDA+, and set $\alpha=1$ and $M=100$ for both gSLDA and gSLDA+.
As we shall see, gSLDA is insensitive to $\alpha$, $c$ and $M$ in a wide range.

Fig.~\ref{fig:20ngBinary} shows the performance of different methods with various numbers of topics. For accuracy, we can draw two conclusions: 1) without making restricting assumptions on the posterior distributions, gSLDA achieves higher accuracy than vSLDA that uses strict variational mean-field approximation; and 2) by using the regularization constant $c$ to improve the influence of supervision information, gSLDA+ achieves much better classification results, in fact comparable with those of MedLDA models since they have the similar mechanism to improve the influence of supervision by tuning a regularization constant. The fact that gLDA+SVM performs better than the standard gSLDA is due to the same reason, since the SVM part of gLDA+SVM can well capture the supervision information to learn a classifier for good prediction, while standard sLDA can't well-balance the influence of supervision. In contrast, the well-balanced gSLDA+ model successfully outperforms the two-stage approach, gLDA+SVM, by performing topic discovery and prediction jointly\footnote{The variational sLDA with a well-tuned $c$ is significantly better than the standard sLDA, but a bit inferior to gSLDA+.}.

For training time, both gSLDA and gSLDA+ are very efficient, e.g., about 2 orders of magnitudes faster than vSLDA and about 1 order of magnitude faster than vMedLDA. For testing time, gSLDA and gSLDA+ are comparable with gMedLDA and the unsupervised gLDA, but faster than the variational vMedLDA and vSLDA, especially when $K$ is large.

\subsection{Multi-class classification}
We perform multi-class classification on the 20NG data set with all the 20 categories.
%The data set has a balanced distribution over the categories. For the test set consisting of 7505 documents, the smallest category has 251 documents and the largest category has $399$ documents. For the training set consisting of 11269 documents, the smallest and the largest categories contain 376 and 599 documents, respectively.
For multi-class classification, one possible extension is to use a multinomial logistic regression model for categorical variables $Y$ by using topic representations $\zvbar$ as input features. However, it is nontrivial to develop a Gibbs sampling algorithm using the similar data augmentation idea, due to the presence of latent variables and the nonlinearity of the soft-max function. In fact, this is harder than the multinomial Bayesian logistic regression, which can be done via a coordinate strategy~\cite{Polson:arXiv12}. Here, we apply the binary gSLDA to do the multi-class classification, following the
%We note that various methods exist to apply a binary classifier to do multi-class classification, such as the ``one-vs-all" and ``one-vs-one" strategies.
``one-vs-all" strategy, which has been shown effective~\cite{Rifkin:jmlr04}, to provide some preliminary analysis. Namely, we learn 20 binary gSLDA models and aggregate their predictions by taking the most likely ones as the final predictions. We again evaluate two versions of gSLDA -- the standard gSLDA with $c=1$ and the improved gSLDA+ with a well-tuned $c$ value. Since gSLDA is also insensitive to $\alpha$ and $c$ for the multi-class task, we set $\alpha=5.6$ for both gSLDA and gSLDA+, and set $c=256$ for gSLDA+. The number of burn-in is set as $M = 40$, which is sufficiently large to get stable results, as we shall see. % in Section~\ref{sec:sensitivity}.

\begin{figure}\vspace{-.1cm}%[h]
\centering
\subfigure[accuracy]{\includegraphics[height=1.1in]{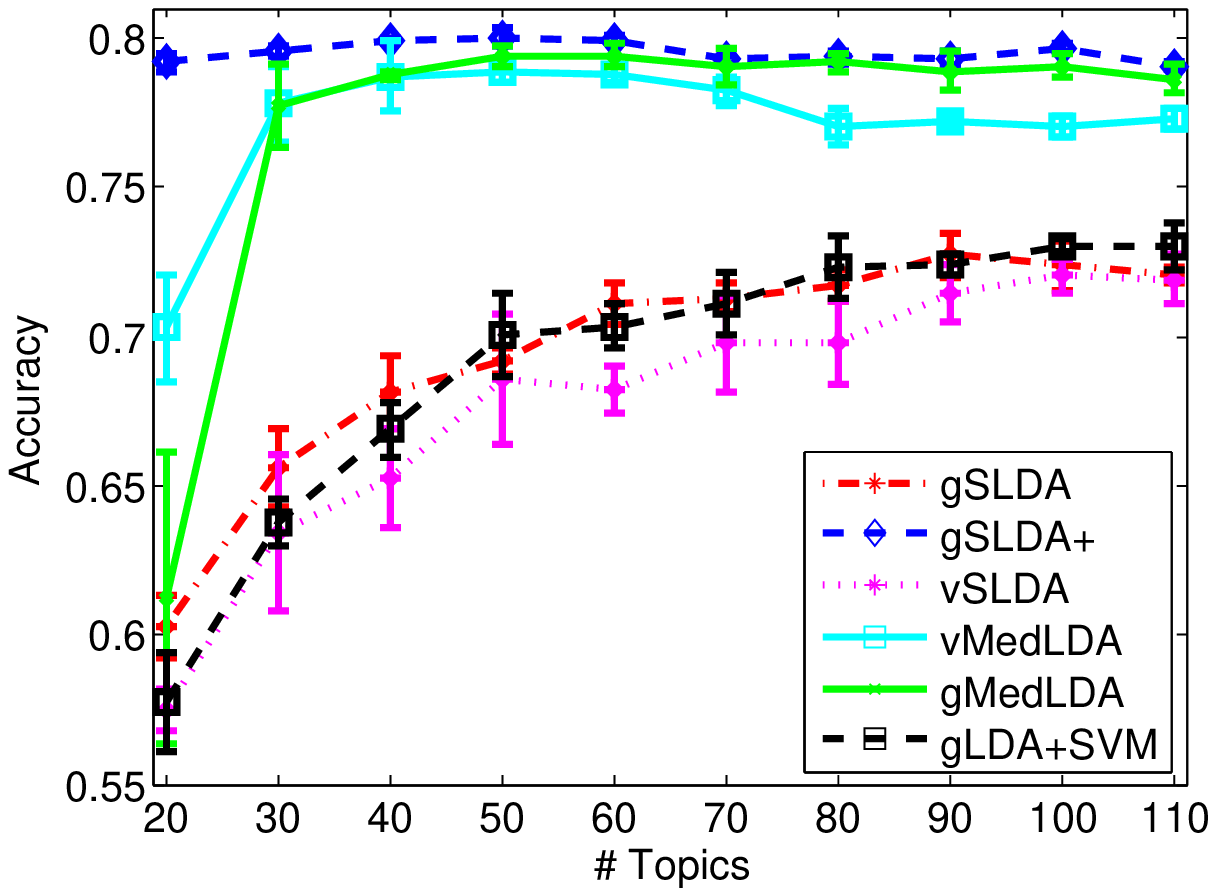}}
\subfigure[training time]{\includegraphics[height=1.1in]{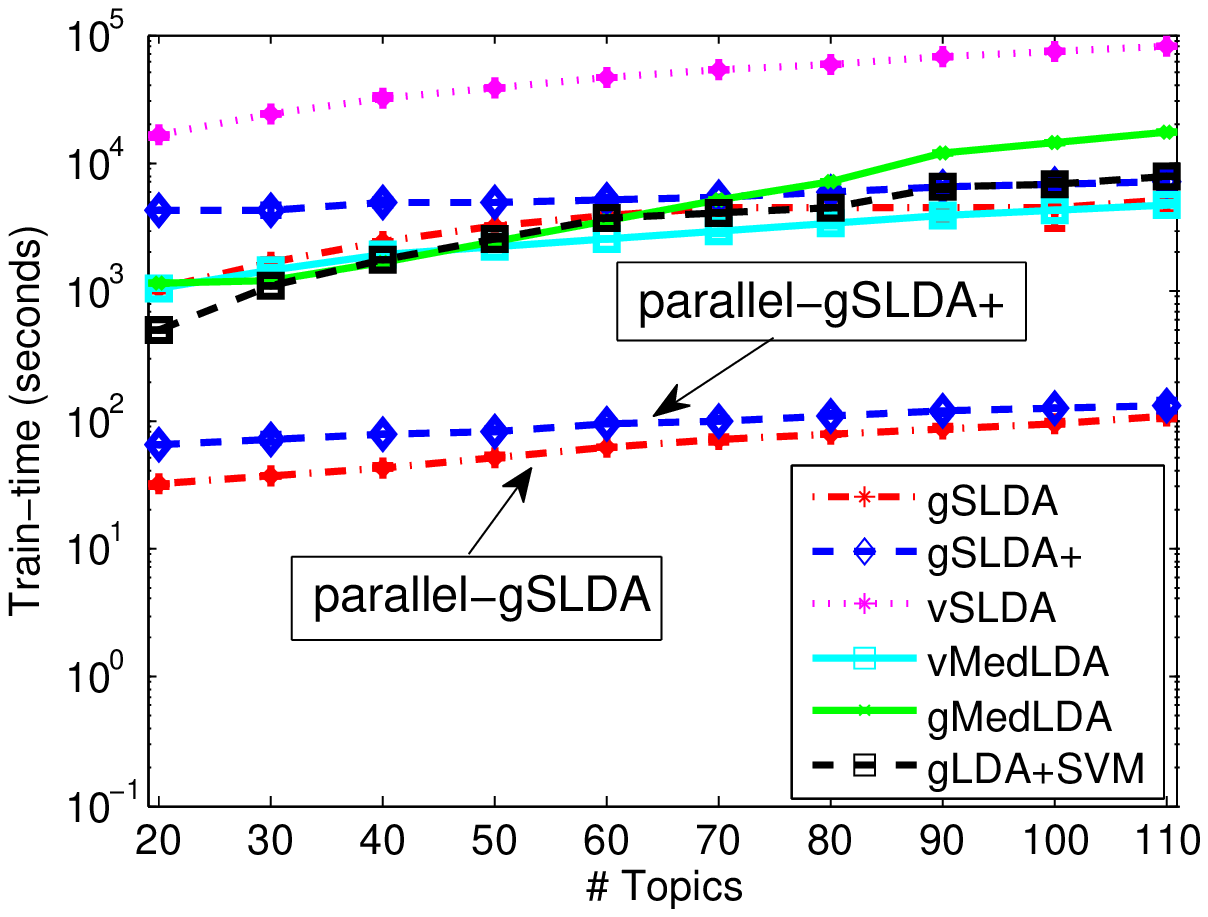}}\vspace{-.3cm}
\caption{Multi-class classification.}
\label{fig:20ng}\vspace{-.4cm}
\end{figure}

Fig.~\ref{fig:20ng} shows the accuracy and training time. We can see that: 1) by using Gibbs sampling without restricting assumptions, gSLDA performs better than the variational vSLDA that uses strict mean-field approximation; 2) due to the imbalance between the single supervision and a large set of word counts, gSLDA doesn't outperform the decoupled approach, gLDA+SVM; and 3) if we increase the value of the regularization constant $c$, supervision information can be better captured to infer predictive topic representations, and gSLDA+ performs much better than gSLDA. In fact, gSLDA+ is even better than the MedLDA that uses mean-field approximation, while is comparable with the MedLDA using collapsed Gibbs sampling. Finally, we should note that the improvement on the accuracy might be due to the different strategies on building the multi-class classifiers. But given the performance gain in the binary task, we believe that the Gibbs sampling algorithm without factorization assumptions is the main factor for the improved performance.

For training time, gSLDA models are about 10 times faster than variational vSLDA. Table~\ref{table:split} shows in detail the percentages of the training time (see the numbers in brackets) spent at each sampling step for gSLDA+. We can see that: 1) sampling the global variables $\etav$ is very efficient, while sampling local variables $(\lambdav, \Zv)$ are much more expensive; and 2) sampling $\lambdav$ is relatively stable as $K$ increases, while sampling $\Zv$ takes more time as $K$ becomes larger. But, the good news is that our Gibbs sampling algorithm can be easily parallelized to speedup the sampling of local variables, following the similar architectures as in LDA.

{\bf A Parallel Implementation}: GraphLab is a graph-based programming framework for parallel computing~\cite{Gonzalez+al:osdi2012}. It provides a high-level abstraction of parallel tasks by expressing data dependencies with a distributed graph. GraphLab implements a GAS (gather, apply, scatter) model, where the data required to compute a vertex (edge) are gathered along its neighboring components, and modification of a vertex (edge) will trigger its adjacent components to recompute their values. Since GAS has been successfully applied to several machine learning algorithms\footnote{http://docs.graphlab.org/toolkits.html} including Gibbs sampling of LDA, we choose it as a preliminary attempt to parallelize our Gibbs sampling algorithm. A systematical investigation of the parallel computation with various architectures in interesting, but beyond the scope of this paper.

\setlength\tabcolsep{3pt}\begin{table}[t]%\vspace{-.3cm}
\caption{Split of training time over various steps.}
\label{table:split}%\vspace{-.2cm}
%\vskip 0.1in
\begin{center}
\begin{small}
\begin{sc}
\scalebox{0.84}{
\begin{tabular}{c|ccc}
\hline
%\abovespace\belowspace
{} & Sample $\lambdav$ & Sample $\etav$ & Sample $\Zv$  \\ %& Total \\
%{} & (\%) & (\%) & (\%) & (seconds) \\
\hline
%\abovespace
%K=20     & 1558.56 (65.80) & 4.26 (0.18) & 805.81 (34.02) & 2368.63  \\ for burn-in = 40
%K=30     & 1547.01 (56.10) & 6.62 (0.24) & 1203.97 (43.66) & 2757.60  \\
%K=40     & 1491.96 (49.00) & 9.13 (0.30) & 1543.72 (50.70) & 3044.82  \\
%K=50     & 1463.65 (43.67) & 11.06 (0.33) & 1876.91 (56.00) & 3351.62  \\
K=20     & 2841.67 (65.80\%) & 7.70 (0.18\%) & 1455.25 (34.02\%) \\% & 4277.62  \\
K=30     & 2417.95 (56.10\%) & 10.34 (0.24\%) & 1888.78 (43.66\%) \\% & 4310.07  \\
K=40     & 2393.77 (49.00\%) & 14.66 (0.30\%) & 2476.82 (50.70\%) \\% & 4885.24  \\
K=50     & 2161.09 (43.67\%) & 16.33 (0.33\%) & 2771.26 (56.00\%) \\% & 4948.68  \\
\hline
\end{tabular}}
\end{sc}
\end{small}
\end{center}\vspace{-.5cm}
\end{table}

For our task, since there is no coupling among the 20 binary gSLDA classifiers, we can learn them in parallel. This suggests an efficient hybrid multi-core/multi-machine implementation, which can avoid the time consumption of IPC (i.e., inter-process communication). Namely, we run our experiments on a cluster with 20 nodes where each node is equipped with two 6-core CPUs (2.93GHz). Each node is responsible for learning one binary gSLDA classifier with a parallel implementation on its 12-cores. For each binary gSLDA model, we construct a bipartite graph connecting train documents with corresponding terms. The graph works as follows: 1) the edges contain the token counts and topic assignments; 2) the vertices contain individual topic counts and the augmented variables $\lambdav$; 3) the global topic counts and $\etav$ are aggregated from the vertices periodically, and the topic assignments and $\lambdav$ are sampled asynchronously during the GAS phases. Once started, sampling and signaling will propagate over the graph. One thing to note is that since we cannot directly measure the number of iterations of an asynchronous model, here we estimate it with the total number of topic samplings, which is again aggregated periodically, divided by the number of tokens.
%For the ``one-vs-all" strategy, since there is no coupling among these classifiers, we can learn the 20 binary gSLDA classifiers in parallel. For the GraphLab implementation, we run 20 instances of the binary classifiers, and one of the instances is selected as the master node. After local binary prediction, the master collects scores of all other nodes and {\bf picks the highest ones??}.
We denote the parallel models by parallel-gSLDA ($c=1$) and parallel-gSLDA+ ($c=256$). From Fig.~\ref{fig:20ng} (b), we can see that the parallel gSLDA models are about 2 orders of magnitudes faster than their sequential counterpart models, which is very promising. Also, the prediction performance is not sacrificed as we shall see in Fig.~\ref{fig:BurnIn-20ng}.
%We also noted that a naive parallelization that learns the 20 binary gSLDA models in parallel with one core per model is about 5 times slower than GraphLab.

\subsection{Sensitivity analysis}\label{sec:sensitivity}
%We now provide more sensitivity analysis of gSLDA to some key parameters.

\begin{figure}\vspace{-.1cm}%[h]
\centering
\subfigure[accuracy]{\includegraphics[height=1in]{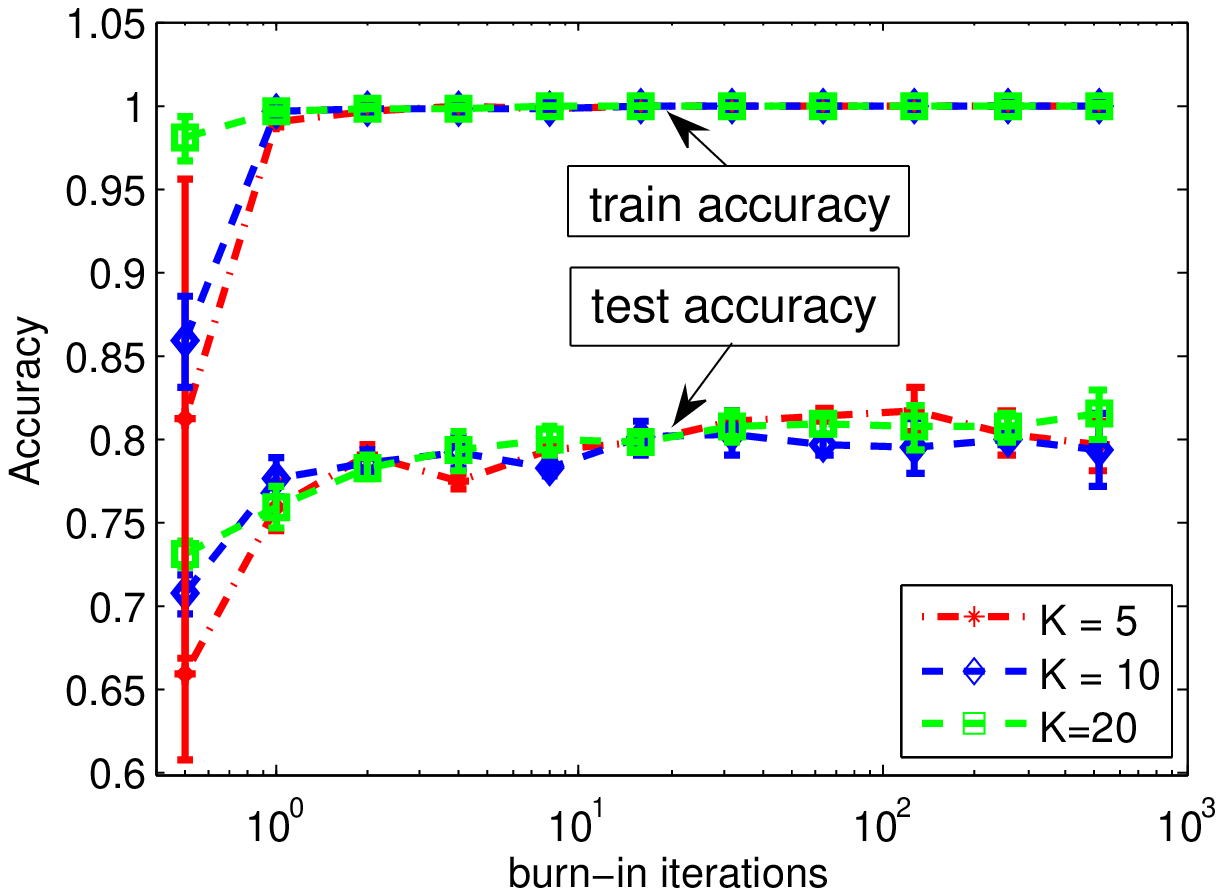}}
\subfigure[training time]{\includegraphics[height=1in]{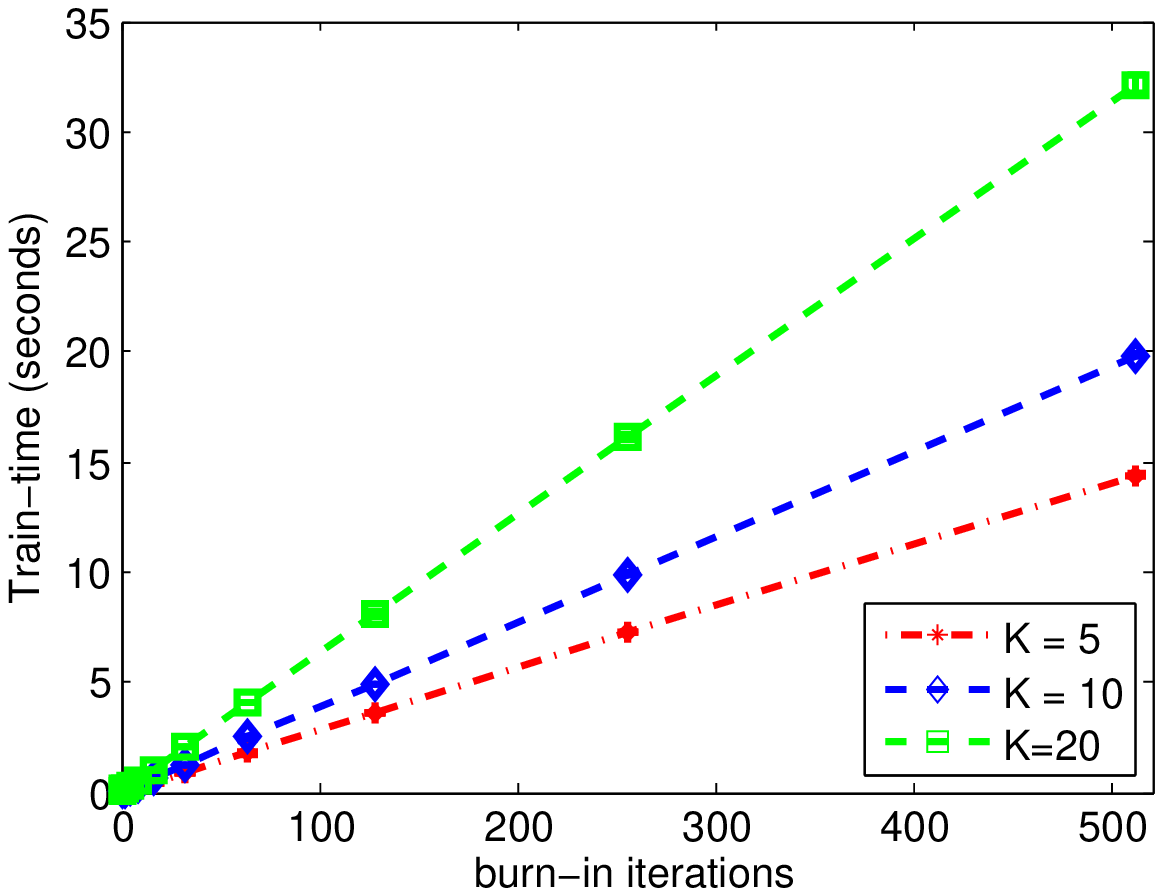}}\vspace{-.3cm}
%\subfigure[]{\includegraphics[height=1.2in]{../fig/acc_binary_20ng_burn-in_v2.eps}}
%\subfigure[]{\includegraphics[height=1.2in]{../fig/train_time_binary_20ng_burn-in_v2.eps}}
\caption{Performance of gSLDA+ with different burn-in steps for binary classification. The most left points are for the settings with no burn in.}
\label{fig:BurnIn-1sample}
\end{figure}

{\bf Burn-In}: Fig.~\ref{fig:BurnIn-1sample} shows the performance of gSLDA+ with different burn-in steps for binary classification. When $M=0$ (see the most left points), the models are built on random topic assignments. We can see that the classification performance increases fast and converges to the stable optimum with about 20 burn-in steps. The training time increases about linearly in general when using more burn-in steps. Moreover, the training time increases linearly as $K$ increases. In the previous experiments, we set $M=100$.
\iffalse
\begin{figure}%[h]
\centering
\subfigure[accuracy]{\includegraphics[height=1.1in]{fig/sensitivity_burn-in_acc_10samples_binary.eps}}
\subfigure[training time]{\includegraphics[height=1.1in]{fig/sensitivity_burn-in_train-time_10samples_binary.eps}}\vspace{-.3cm}
%\subfigure[]{\includegraphics[height=1.2in]{../fig/acc_binary_20ng_burn-in_v2.eps}}
%\subfigure[]{\includegraphics[height=1.2in]{../fig/train_time_binary_20ng_burn-in_v2.eps}}
\caption{Classification accuracy and training time of gSLDA+ with different burn-in steps for binary classification. The final classifier is an average of 10 sampled classifiers.}
\label{fig:BurnIn}
\end{figure}
\fi
\begin{figure}%[h]
\centering
\subfigure[accuracy]{\includegraphics[height=1.1in]{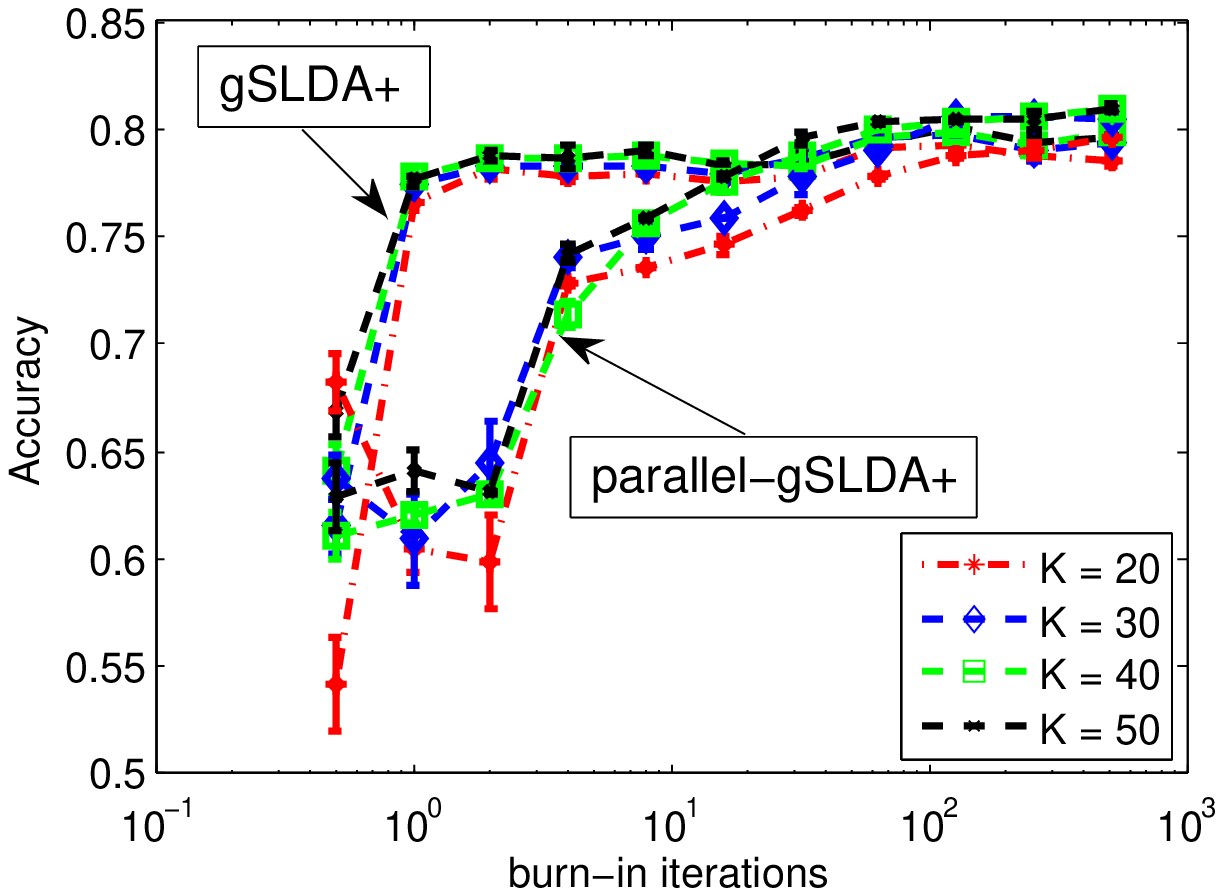}}
\subfigure[training time]{\includegraphics[height=1.1in]{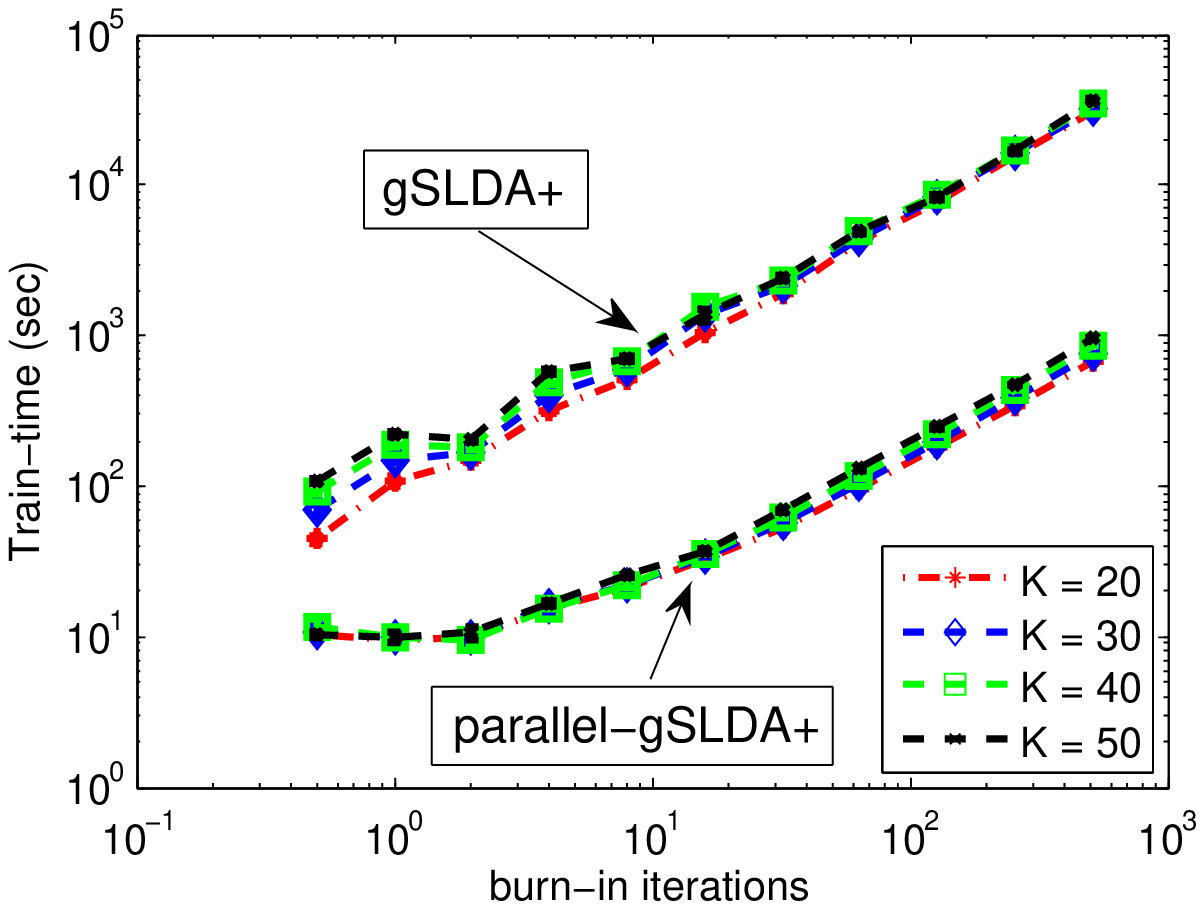}}\vspace{-.2cm}
%\subfigure[]{\includegraphics[height=1.2in]{../fig/acc_binary_20ng_burn-in_v2.eps}}
%\subfigure[]{\includegraphics[height=1.2in]{../fig/train_time_binary_20ng_burn-in_v2.eps}}
\caption{Performance of gSLDA+ and parallel-gSLDA+ with different burn-in steps for multi-class classification. The most left points are for the settings with no burn in.}\vspace{-.3cm}
\label{fig:BurnIn-20ng}
\end{figure}

Fig.~\ref{fig:BurnIn-20ng} shows the performance of gSLDA+ and its parallel implementation (i.e., parallel-gSLDA+) for the multi-class classification with different burn-in steps. We can see when the number of burn-in steps is larger than 20, the performance of gSLDA+ is quite stable. Again, in the log-log scale, since the slopes of the lines in Fig.~\ref{fig:BurnIn-20ng} (b) are close to the constant 1, the training time grows about linearly as the number of burn-in steps increases. Even when we use 40 or 60 burn-in steps, the training time is still competitive, compared with the variational vSLDA. For parallel-gSLDA+ using GraphLab, the training is consistently about 2 orders of magnitudes faster. Meanwhile, the classification performance is also comparable with that of gSLDA+, when the number of burn-in steps is larger than 40. In the previous experiments, we have set $M=40$ for both gSLDA+ and parallel-gSLDA+.

%We also observed that the naive parallel implementation that learns the 20 binary gSLDA+ models in parallel is about 3 times slower than the parallel-gSLDA+ with GraphLab.

% is not sa, especially considering that gSLDA can be naively parallelized by learning different binary classifiers simultaneously. In the previous experiments, we have set $M=64$.
\iffalse
\begin{figure}%[h]
\centering
\includegraphics[height=1.7in]{../fig/20ng_burnIn.eps}\vspace{-.3cm}
%\subfigure[]{\includegraphics[height=1.4in]{../fig/acc_20ng_burn-in.eps}}
%\subfigure[]{\includegraphics[height=1.4in]{../fig/train_time_20ng_burn-in.eps}}
%\subfigure[]{\includegraphics[height=1.4in]{../fig/train_time_20ng_burn-in_parallel.eps}}
\caption{(left) classification accuracy, (middle) training time of GibbsMedLDA, and (right) training time of pGibbsMedLDA with different numbers of burn-in steps for multi-class classification.}
\label{fig:BurnIn-20ng}
\end{figure}
\fi

{\bf Regularization constant $c$}: Fig.~\ref{fig:Sensitivity-Ell} shows the performance of gSLDA in the binary classification task with different $c$ values. We can see that in a wide range, e.g., from 9 to 100, the performance is quite stable for all the three $K$ values. But for the standard sLDA model, i.e., $c=1$, both the training accuracy and test accuracy are low, which indicates that sLDA doesn't fit the supervision data well. When $c$ becomes larger, the training accuracy gets higher, but it doesn't seem to over-fit and the generalization performance is stable. In the above experiments, we set $c = 25$. For multi-class classification, we have similar observations and set $c=256$ in the previous experiments.
\begin{figure}%[h]
\centering
\subfigure[accuracy]{\includegraphics[height=1.1in]{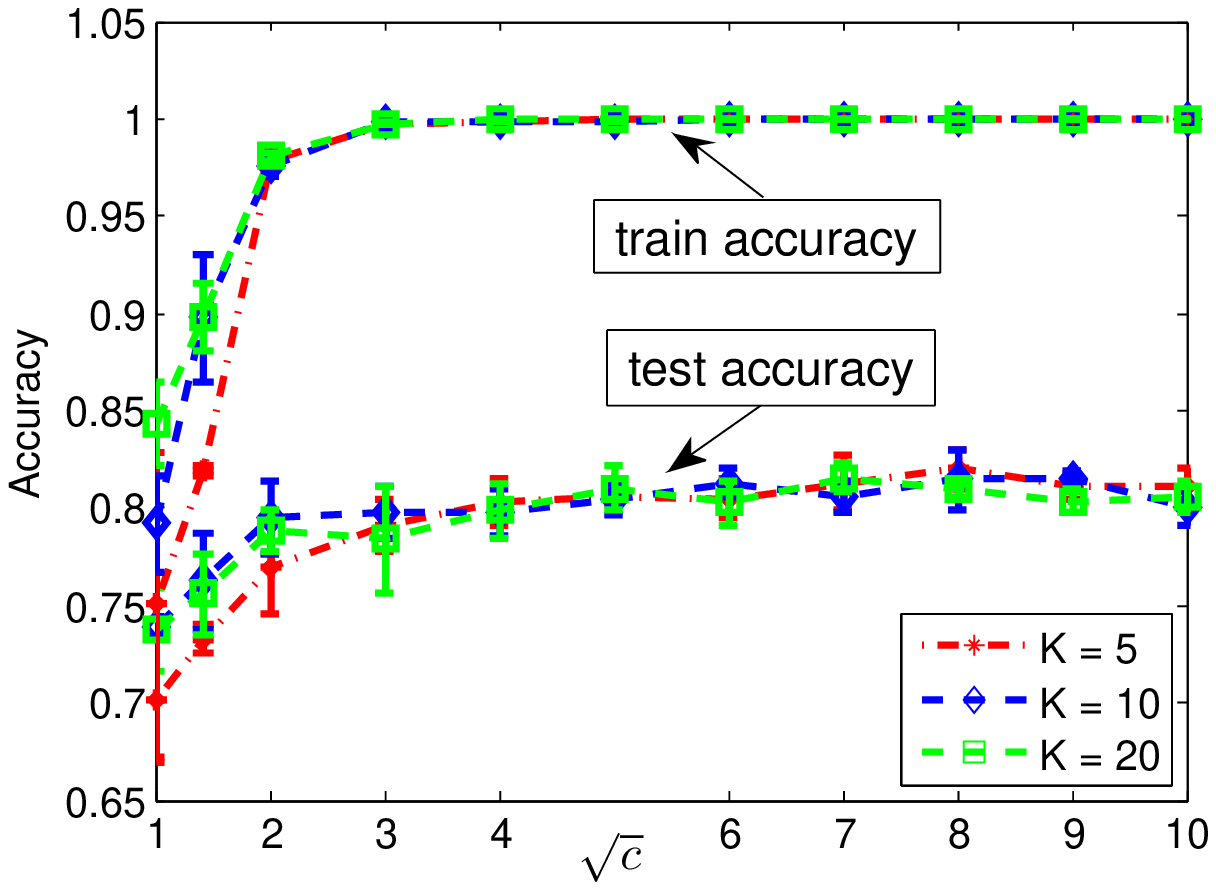}}
\subfigure[training time]{\includegraphics[height=1.1in]{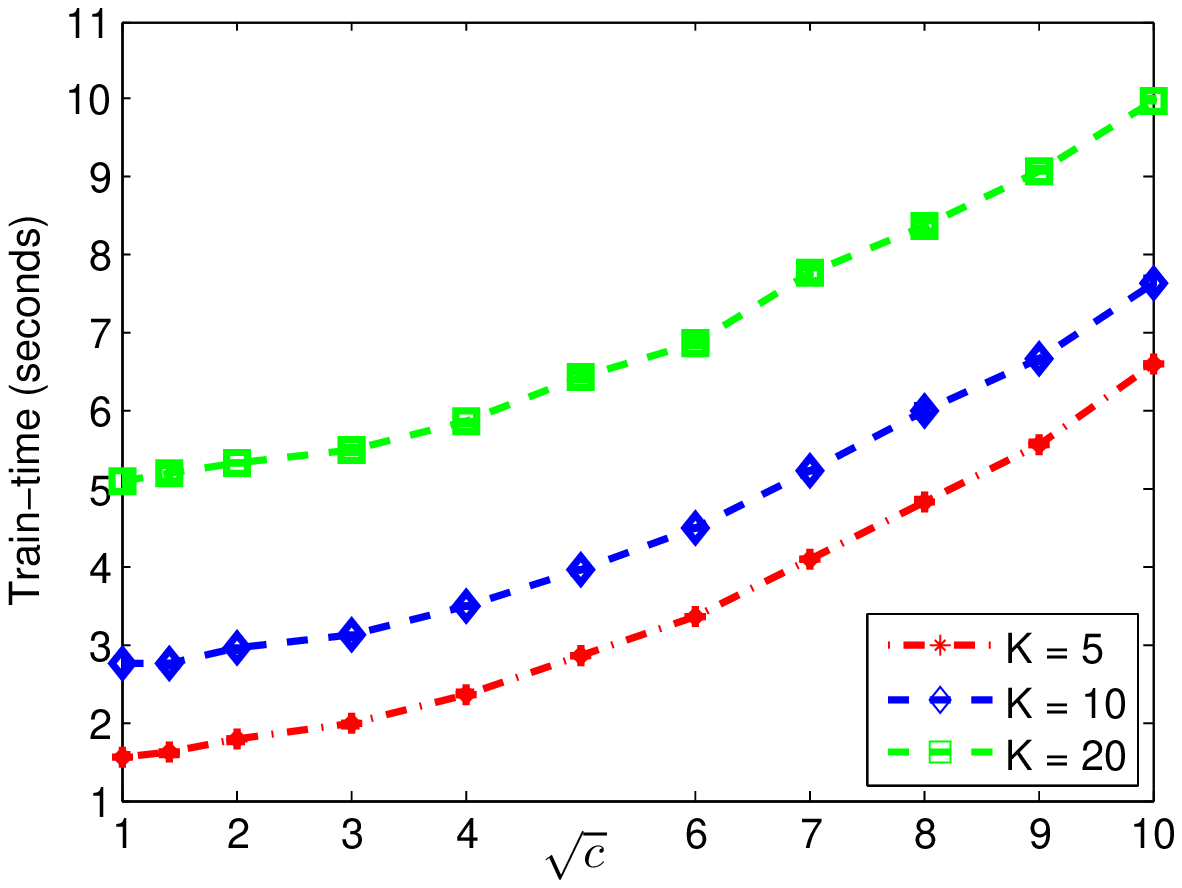}}\vspace{-.3cm}
\caption{Performance of gSLDA for binary classification with different $c$ values.}\label{fig:Sensitivity-Ell}\vspace{-.2cm}
\end{figure}
\iffalse
\begin{figure}%[h]
\centering
\subfigure[accuracy]{\includegraphics[height=1.1in]{fig/sensitivity_c_acc_binary.eps}}
\subfigure[training time]{\includegraphics[height=1.1in]{fig/sensitivity_c_train-time_binary.eps}}\vspace{-.3cm}
\caption{Performance of sLDA on the binary classification data set with different $c$ values.}\label{fig:Sensitivity-Ell}\vspace{-.2cm}
\end{figure}
\fi

\begin{figure}%[h]
\centering
\subfigure[$c=1$]{\includegraphics[height=1in]{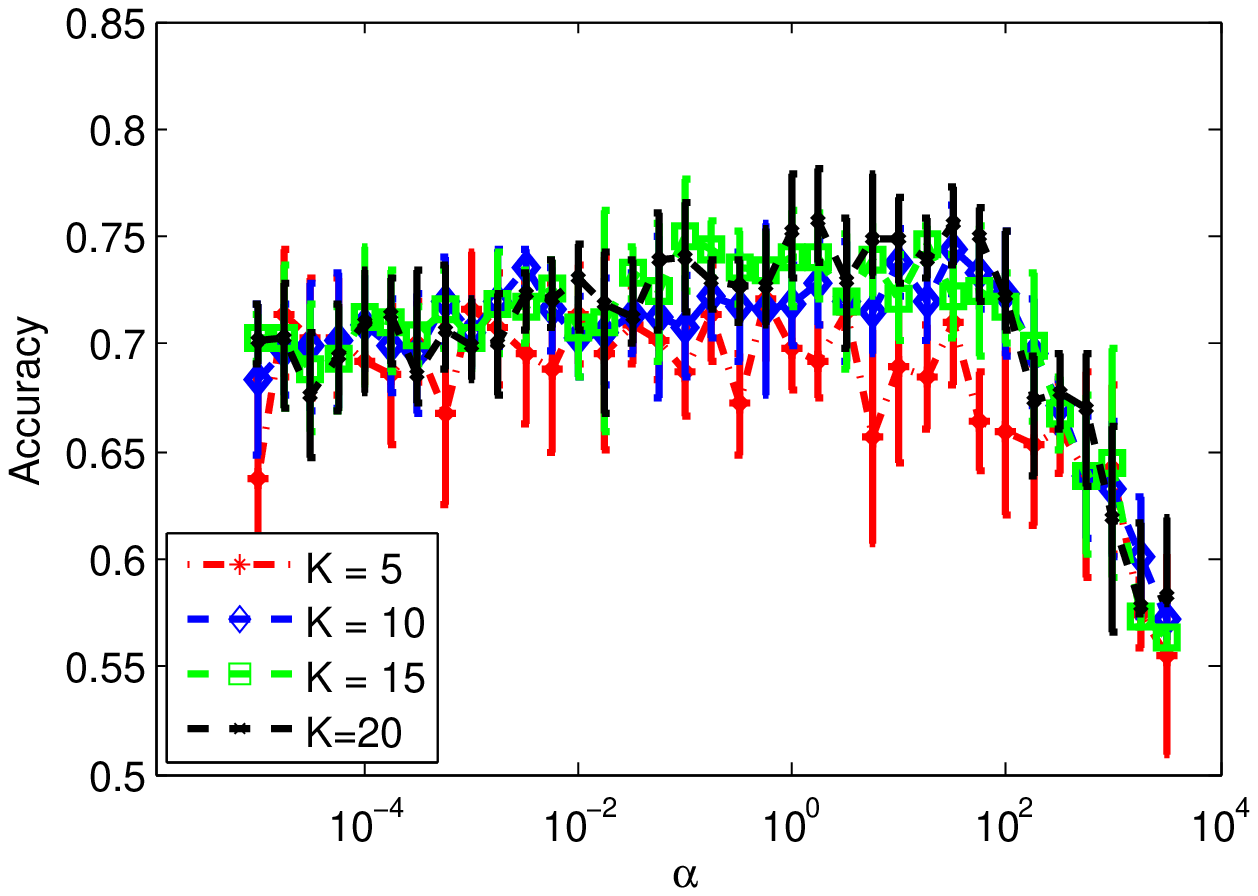}}
\subfigure[$c=9$]{\includegraphics[height=1in]{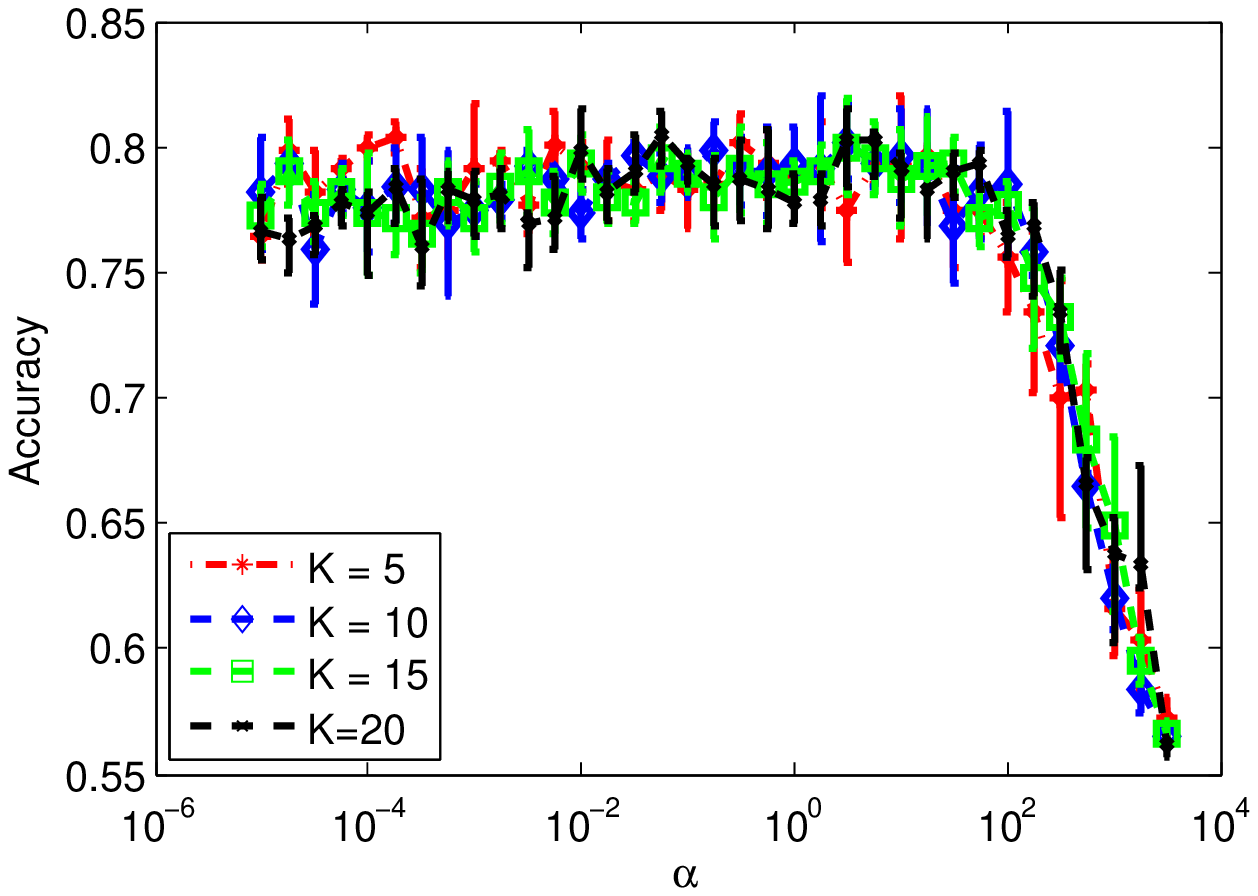}}\vspace{-.3cm}
\caption{Accuracy of gSLDA for binary classification with different $\alpha$ values in two settings with $c=1$ and $c=9$.}\label{fig:Sensitivity-Alpha}\vspace{-.2cm}
\end{figure}

{\bf Dirichlet prior $\alpha$}:
Fig.~\ref{fig:Sensitivity-Alpha} shows the performance of gSLDA on the binary task with different $\alpha$ values. We report two cases with $c=1$ and $c=9$. %For the four different topic numbers,
We can see that the performance is quite stable in a wide range of $\alpha$ values, e.g., from $0.1$ to $10$. We also noted that the change of $\alpha$ does not affect the training time much.
%We can also see that it generally needs a larger $\alpha$ in order to get stable results when $K$ becomes larger. This is mainly because a large $K$ tends to produce sparse topic representations and an appropriately large $\alpha$ is needed to smooth the representations, as the effective Dirichlet prior is $\alpha_k = \alpha / K$.

\section{Conclusions and Discussions}\label{section:conclusions}

We present two improvements to Bayesian logistic supervised topic models, namely, a general formulation by introducing a regularization parameter to avoid model imbalance and a highly efficient Gibbs sampling algorithm without restricting assumptions on the posterior distributions by exploring the idea of data augmentation. The algorithm can also be parallelized. Empirical results for both binary and multi-class classification demonstrate significant improvements over the existing logistic supervised topic models. Our preliminary results with GraphLab have shown promise on parallelizing the Gibbs sampling algorithm.

For future work, we plan to carry out more careful investigations, e.g., using various distributed architectures~\cite{Ahmed:WSDM12,Newman08distributedinference,Smola:vldb10}, to make the sampling algorithm highly scalable to deal with massive data corpora. Moreover, the data augmentation technique can be applied to deal with other types of response variables, such as count data with a negative-binomial likelihood~\cite{Polson:arXiv12}.

\section*{Acknowledgments}
This work is supported by National Key Foundation R\&D Projects (No.s 2013CB329403, 2012CB316301), Tsinghua Initiative
Scientific Research Program No.20121088071, Tsinghua National Laboratory for Information Science and
Technology, and the 221 Basic Research Plan for Young Faculties at Tsinghua University.

%Do not number the acknowledgment section. Do not include this section
%when submitting your paper for review.
\bibliographystyle{acl}
% you bib file should really go here
\bibliography{slda}

\end{document}